\documentclass[journal]{IEEEtran}
\usepackage[usenames,dvipsnames]{color}
\usepackage{colortbl}
\usepackage{pifont}       
\usepackage{bbding}       
\usepackage{fontawesome}  
\definecolor{mygray}{gray}{.9}
\usepackage{amsmath}
\usepackage{algorithm}
\usepackage{algorithmic}
\usepackage{graphicx}
\usepackage{verbatim}
\usepackage{multirow}
\usepackage{tabularx}
\usepackage{xcolor}
\usepackage{array}

\usepackage{amssymb}
\usepackage{makecell}
\usepackage{subfigure}
\usepackage{diagbox}
\usepackage{rotating}
\usepackage{booktabs}
\usepackage{tablefootnote}

\definecolor{darkgreen}{rgb}{0.0, 0.5, 0.0}

\usepackage{cite}
\usepackage[colorlinks,
            linkcolor=red,
            anchorcolor=blue,
            citecolor=green
            ]{hyperref}

\newcommand{\addFig}[1]{}
\newcommand{\addFigs}[1]{}

\usepackage{subfigure}

\usepackage{balance}
\usepackage{cleveref}
\crefformat{section}{\S#2#1#3} 
\crefformat{subsection}{\S#2#1#3}
\crefformat{subsubsection}{\S#2#1#3}

\ifCLASSINFOpdf
\else
\fi

\begin{document}
%
\title{Unleashing the Power of Motion and Depth: A Selective Fusion Strategy for RGB-D Video Salient Object Detection}
%
%
%
\author{Jiahao He, Daerji Suolang, Keren Fu, and Qijun Zhao \thanks{J. He, D. Suo, K. Fu and Q. Zhao are with the College of Computer Science, Sichuan University, China. (Email: 1422703074@qq.com; sonam2@163.com; fkrsuper@scu.edu.cn; qjzhao@scu.edu.cn;).\\
\indent Corresponding author: Keren Fu \\
}
}

\markboth{JOURNAL OF LATEX CLASS FILES, VOL. 14, NO. 8, AUGUST 2015}%
{Shell \MakeLowercase{\textit{et al.}}: Bare Demo of IEEEtran.cls for IEEE Journals}

\maketitle

\begin{abstract}
Applying salient object detection (SOD) to RGB-D videos is an emerging task called RGB-D VSOD and has recently gained increasing interest, due to considerable performance gains of incorporating motion and depth and that RGB-D videos can be easily captured now in daily life. 
Existing RGB-D VSOD models have different attempts to derive motion cues, in which extracting motion information explicitly from optical flow appears to be a more effective and promising alternative. Despite this, there remains a key issue that how to effectively utilize optical flow and depth to assist the RGB modality in SOD. Previous methods always treat optical flow and depth equally with respect to model designs, without explicitly considering their unequal contributions in individual scenarios, limiting the potential of motion and depth. 
To address this issue and unleash the power of motion and depth, we propose a novel selective cross-modal fusion framework (SMFNet) for RGB-D VSOD, incorporating a pixel-level selective fusion strategy (PSF) that achieves optimal fusion of optical flow and depth based on their actual contributions. Besides, we propose a multi-dimensional selective attention module (MSAM) to integrate the fused features derived from PSF with the remaining RGB modality at multiple dimensions, effectively enhancing feature representation to generate refined features. 
We conduct comprehensive evaluation of SMFNet against 19 state-of-the-art models on both RDVS and DVisal datasets, making the evaluation the most comprehensive RGB-D VSOD benchmark up to date, and it also demonstrates the superiority of SMFNet over other models. Meanwhile, evaluation on five video benchmark datasets incorporating synthetic depth validates the efficacy of SMFNet as well. Our code and benchmark results are made publicly available at \href{https://github.com/Jia-hao999/SMFNet} {https://github.com/Jia-hao999/SMFNet}.
\end{abstract}


\IEEEpeerreviewmaketitle

\section{INTRODUCTION} \label{sec1}

\IEEEPARstart {S}{alient} object detection (SOD) refers to segmenting out the most visually distinctive objects that capture human attention within a given scene. This task is recognized as an essential component in the field of data processing, commanding increasing attention in recent years. SOD can be applied to a variety of computer vision tasks, including but not limited to semantic segmentation \cite{wei2016stc}, object detection \cite{wu2022single}, image retrieval \cite{cheng2017intelligent} and person re-identification \cite{zhao2013unsupervised}.

\begin{figure}
	\centering
	\includegraphics[width=0.48\textwidth]{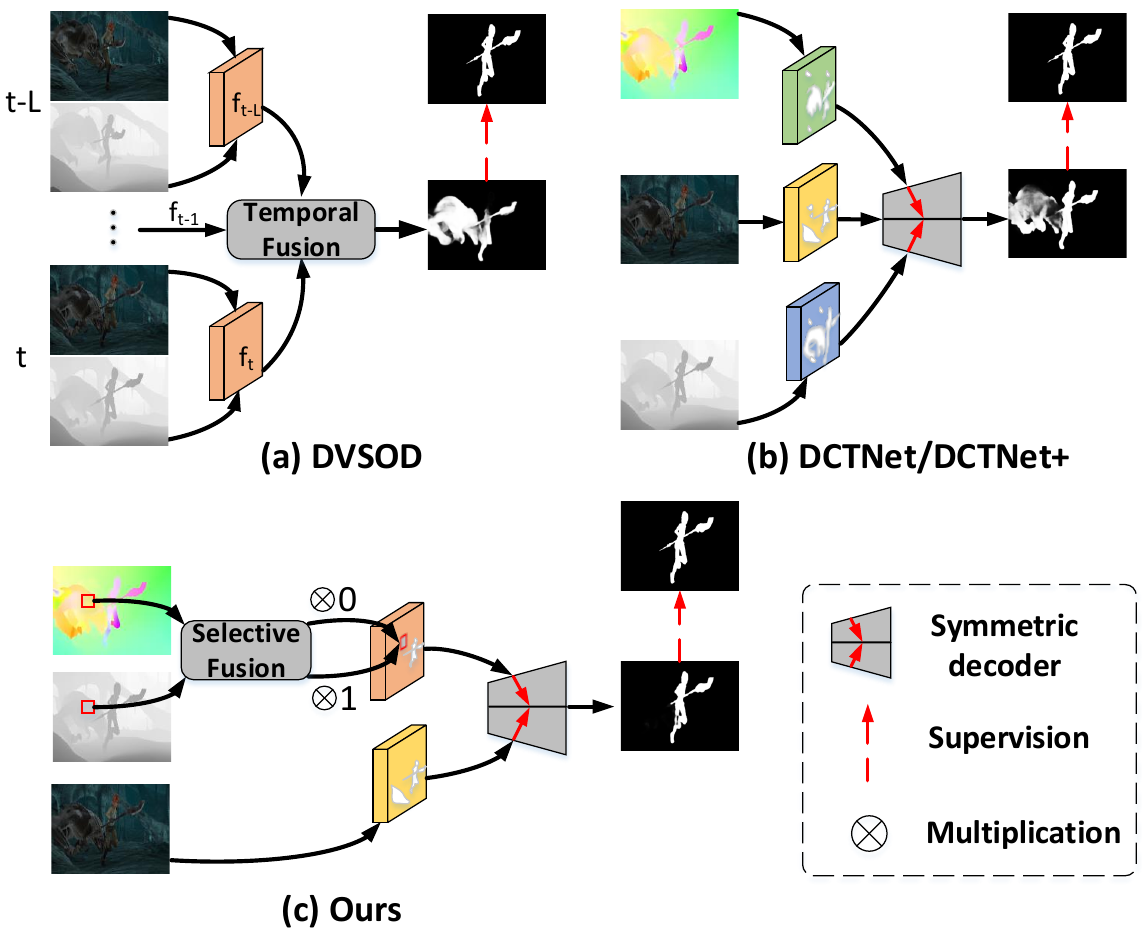}
    \vspace{-0.3cm}
	\caption{\small{Comparison between existing RGB-D VSOD frameworks and the proposed model. (a) DVSOD \cite{li2024dvsod} performs temporal fusion of RGB and depth features from a few frames to derive motion information. (b) DCTNet \cite{lu2022depth} and DCTNet+ \cite{mou2023salient} explicitly derive motion information from optical flow, and feed all extracted features from RGB, optical flow and depth to a symmetric decoder. (c) Our method first conducts pixel-level selective fusion of optical flow and depth, and then feed RGB and the fused features to the decoder.
    \label{fig_introduction}}}
    \vspace{-0.4cm}
\end{figure}

Although SOD based on the single RGB modality has shown remarkable performance in this field, it appears notably restrictive when encountering complex environment. To overcome this daunting challenge, researchers introduce scene depth information into the SOD task, yielding an emerging field called RGB-D salient object detection (RGB-D SOD). Meanwhile, as most real-world scenes are dynamic, researchers also extends the SOD task to a dynamic setup called video salient object detection (VSOD). Although RGB-D SOD and VSOD have been extensively studied and advanced by researchers in the past decade, applying SOD to RGB-D videos (i.e., the RGB-D VSOD task) is still in a preliminary stage of exploration \cite{lu2022depth,mou2023salient,li2024dvsod}. 

Despite that there are some previous works to compute saliency in 3D stereoscopic videos \cite{zhang2010stereoscopic,ferreira2015method,zhang2020stereoscopic,kim2014saliency}, they are only based on traditional computational methods and do not involve deep learning. On the other hand, limited by the lack of RGB-D video datasets in this field,  RGB-D VSOD researches based on convolutional neural networks (CNNs) have not yet been widely investigated. 
Thanks to the success of monocular depth estimation technique, Lu et al. \cite{lu2022depth} firstly use synthetic depth as an alternative to assist VSOD and achieve encouraging improvements on VSOD benchmarks. However, synthetic depth maps are sometimes hard to reflect real-world depth information, so this work \cite{lu2022depth} has certain limitations in practice. Fortunately, the extended research by Mou et al. \cite{mou2023salient} based on this work comes up, in which an RGB-D video dataset named RDVS with realistic depth maps is proposed. Concurrently, a larger and more comprehensive RGB-D video dataset called DVisal is proposed by Li et al \cite{li2024dvsod}, which not only provides data foundation for the community, but also demonstrates the significance of exploring this field. It is worth mentioning that the two works above also construct methods/models for RGB-D VSOD, to validate the usefulness of incorporating depth and provide potential directions for subsequent study.

Fig. \ref{fig_introduction} (a) and (b) illustrate two architectures of the existing RGB-D VSOD models \cite{lu2022depth, mou2023salient, li2024dvsod}. One can see that the most significant difference between them lies in how to derive motion information. DVSOD \cite{li2024dvsod} performs temporal fusion over several frames to implicitly model motion stimuli. However, the fusion of multiple frames may inevitably bring redundancy and noise, which seriously interferes with final prediction. In contrast, DCTNet \cite{lu2022depth} and DCTNet+ \cite{mou2023salient} first derive optical flow between two adjacent frames, and from the optical flow, motion information can be explicitly extracted. Although an extra step is required to compute the optical flow, the extracted motion information is more dedicated to final prediction. Although DCTNet and DCTNet+ have made encouraging progresses, there remains a key issue not considered, limiting the potential of incorporating optical flow and depth in this task. That is, DCTNet and DCTNet+ adopt symmetric decoders to fuse optical flow and depth into the RGB branch (Fig. \ref{fig_introduction} (b)), which implies equal contributions for optical flow and depth with respect to model designs. However, since optical flow and depth each cannot always be robust and useful across all scenarios, they naturally hold unequal values in different scenarios, and therefore effectively utilizing optical flow and depth requires considering their actual contributions.

To address this dilemma and unleash the power of motion and depth, we propose a novel selective cross-modal fusion framework (SMFNet), as shown in Fig. \ref{fig_introduction} (c). We still keep the trimodal input fashion (i.e., RGB, optical flow, and depth) to extract dedicated features. However, unlike DCTNet and DCTNet+ \cite{lu2022depth, mou2023salient}, we design a pixel-level selective fusion strategy (PSF) to selectively fuse optical flow and depth before the decoding process. 
Specifically, we first integrate all extracted features of optical flow and depth to derive a spatial weight map, which is then used to compute the weighed sum of these features. A pseudo-supervisory algorithm is employed to generate pseudo ground truth based on contributions of optical flow and depth. During training, we use such pseudo ground truth to supervise the spatial weight map, in order to guarantee its efficacy and correctness.
In the subsequent symmetric decoder, we propose a multi-dimensional selective attention module (MSAM) to fully promote cross-modal interactions. 
Given RGB features and also the fused features derived from PSF, MSAM conducts selective attention on width, height, spatial and channel dimensions, respectively, to enhance feature representation and generate refined features.

In a nutshell, this paper provides three main contributions:
\begin{itemize}
	\item
  We propose a pixel-level selective fusion strategy (PSF), incorporating the process of generating a spatial weight map and its pseudo-supervisory algorithm. PSF selects and fuses the most valuable features of optical flow and depth pixel-by-pixel based on their actual contributions.
  
	\item
  We propose a multi-dimensional selective attention module (MSAM) to integrate cross-modal features at multiple dimensions. It can generate multiple attention weights through cross-modal perceptual interactions, thus effectively enhancing the representation of integrated features.

    \item
  The proposed SMFNet, equipped with PSF and MSAM, is the first RGB-D VSOD model to be evaluated on both RDVS and DVisal datasets. Thus, we conduct comprehensive evaluation of 19 state-of-the-art (SOTA) models together with SMFNet on RDVS and DVisal, making the evaluation the most comprehensive RGB-D VSOD benchmark up to date. Extensive experiments conclusively demonstrate that SMFNet outperforms SOTA models. 
  Besides, we evaluate SMFNet on VSOD benchmark datasets equipped with synthetic depth maps. The experimental results also show the superiority of SMFNet.
  The benchmark results are made available 
  at \href{https://github.com/Jia-hao999/SMFNet} {https://github.com/Jia-hao999/SMFNet}.
  
\end{itemize}

\section{RELATED WORK}\label{Related_Works}
\subsection{RGB-D Salient Object Detection}\label{sec21}
In early days, RGB-D SOD works \cite{peng2014rgbd,cong2016saliency} tended to extract hand-crafted features and then fused RGB images and depth maps. However, these methods are difficult to extract effective features in complex scenes, resulting in great limitations. Thanks to the vigorous development of deep learning, CNN-based RGB-D SOD models \cite{fu2020jl,ji2021calibrated,zhou2021specificity,cheng2022depth,chen20223,hu2024cross} have gradually become the mainstream and achieve superiority performance. 
Fu et al. \cite{fu2020jl} proposed a joint learning and densely-cooperative fusion framework (JL-DCF) to effectively extract and fuse deep hierarchical features from RGB and depth inputs. To reduce the negative effects of inaccurate depth maps, Ji et al. \cite{ji2021calibrated} designed a depth calibration strategy (DC) to calibrate the depth images. To explore the shared information as well as preserve
modality-specific characteristics, Zhou et al. \cite{zhou2021specificity} proposed a novel specificity-preserving network (SPNet) for RGB-D SOD. Recently, a cross-modal fusion and progressive decoding network (CPNet) is proposed by Hu et al. \cite{hu2024cross}, to effectively carry out multi-scale feature aggregation. Because the ability of CNNs in learning global contexts is limited, Transformers are then widely introduced in recent RGB-D SOD works \cite{tang2022hrtransnet,sun2023catnet,cong2023point,guo2024unitr} to bridge this gap. Tang et al. \cite{tang2022hrtransnet} proposed a unified two-modality SOD model (HRTransNet) to maintain high-resolution representation with a large receptive field. Sun et al. \cite{sun2023catnet} proposed a cascaded and aggregated Transformer network (CATNet), which adopts Swin Transformer as the backbone network to extract global semantic information of RGB and depth.

\subsection{Video Salient Object Detection}\label{sec22}
In the field of VSOD, traditional methods mainly rely on hand-crafted features and prior knowledge, such as color-contrast \cite{li2015spatiotemporal}, background prior \cite{han2014background} and morphology cues \cite{rahtu2010segmenting}. But the performance of these approaches is limited by the representation ability of low level features. Subsequently, the emergence of deep learning-based methods breaks this limitation and continues to advance the detection performance. Wang et al. \cite{wang2017video} proposed the first model for applying deep learning to VSOD, which is much faster than traditional video saliency models in dynamic scenes. Li et al. \cite{li2018flow} introduced a flow guided recurrent neural encoder framework to enhance the temporal coherence modeling of the per-frame feature representation. Subsequent work \cite{li2019motion} also introduced optical flow into the model, but it took optical flow as a separate branch for feature extraction and fused optical flow features with RGB features to explicitly capture motion information. Zhang et al. \cite{zhang2021dynamic} proposed a dynamic context-sensitive filtering module to estimate the location-related
affinity weights to dynamically generate context sensitive convolution kernels. Due to that existing data-driven approaches heavily rely on a large quantity of pixel-wise annotated video frames, Piao et al. \cite{piao2022semi} proposed a pseudo label generator, which can make full use of inter-frame information to locate salient objects in unlabeled frames.

\subsection{RGB-D Video Salient Object Detection}\label{sec23}
Our investigation shows that there are very few researches on RGB-D VSOD at present because of the lack of RGB-D video datasets that are suitable for this task. However, there are still some preliminary works. To simulate human visual system in the 3D world, Zhang et al. \cite{zhang2010stereoscopic} first proposed a bottom-up Stereoscopic Visual Attention (SVA) model, integrating depth, appearance and motion information to detect the most attractive objects. Considering different contributions of multi-source information to saliency, Kim et al. \cite{kim2014saliency} calculated saliency intensity of motion, depth and appearance attributes, respectively, and then fused the resulting saliency maps based on such saliency intensity. Lino et al. \cite{ferreira2015method} proposed a computational method to determine saliency regions in 3D videos, based on fusion of three feature maps containing perceptually relevant cues from spatial, temporal and depth dimensions. 

The above methods are all based on traditional saliency computation, and do not involve deep learning, so their detection performance is limited. To explore the contribution of depth in VSOD, Lu et al. \cite{lu2022depth} proposed a depth-cooperated trimodal network (DCTNet), 
in which optical flow and depth features enhance RGB features to promote detection performance. However, due to the lack of suitable benchmark datasets, their method utilized synthetic depth maps, instead of realistic depth maps. Later, based on the previous work \cite{lu2022depth}, Mou et al. \cite{mou2023salient} constructed an RDVS dataset containing realistic depth maps and also proposed an improved trimodal network called DCTNet+. Compared with DCTNet, DCTNet+ achieves notable performance improvement. More recently, Li et al. \cite{li2024dvsod} proposed another comprehensively annotated RGB-D video dataset named DViSal, providing further support for research in this field. In \cite{li2024dvsod}, an RGB-D VSOD baseline model is also introduced to demonstrate the advantages of incorporating depth information into videos for SOD.

\section{PROPOSED METHOD}\label{The_Proposed_Model}
\subsection{Overview}\label{sec31}

\begin{figure*}
	\centering
	\includegraphics[width=0.98\linewidth,height=330pt]{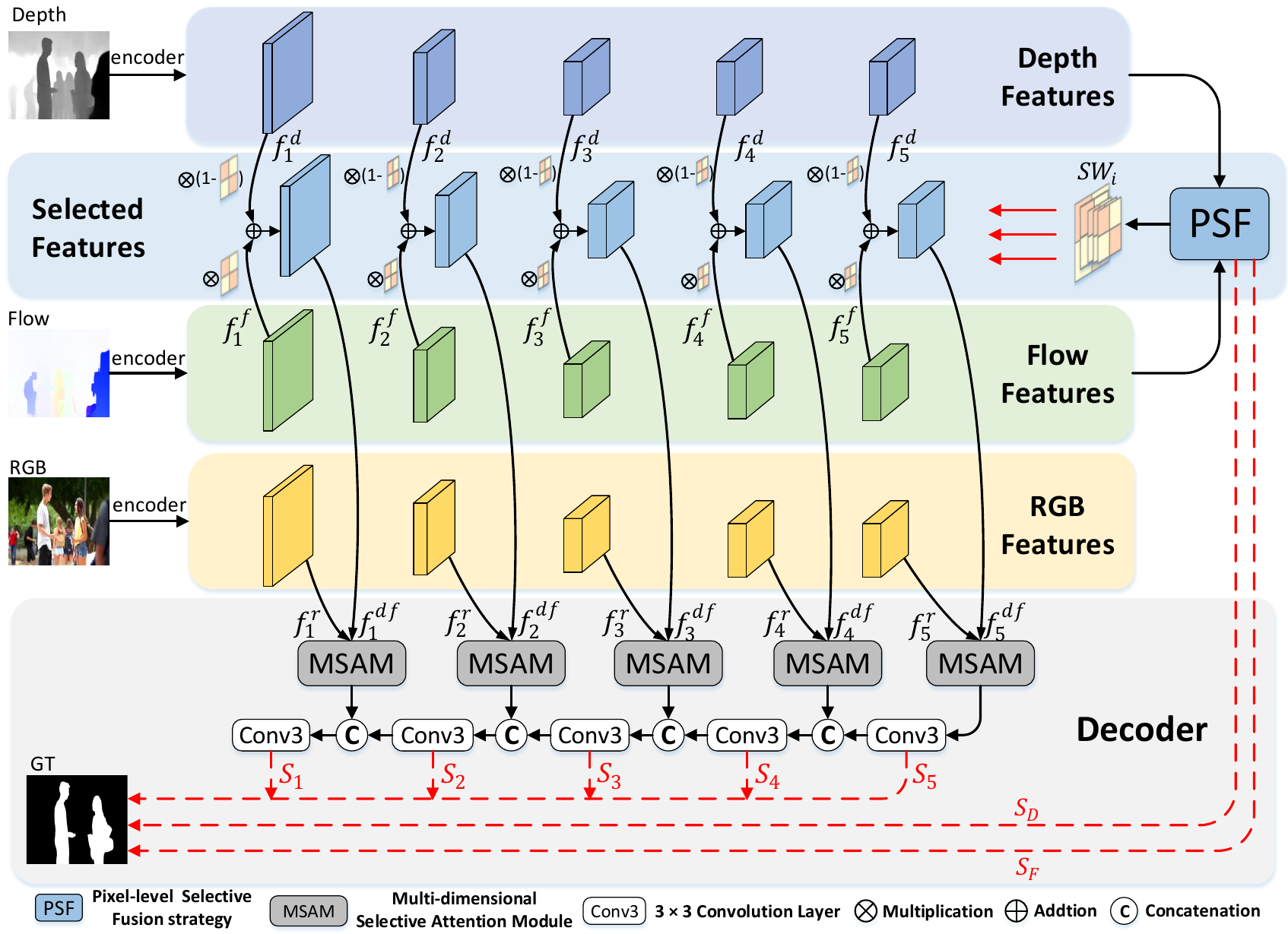}
    \vspace{-0.3cm}
	\caption{\small {Overview of the proposed SMFNet.}}
    \label{fig_overview}
    \vspace{-0.4cm}
\end{figure*}

\begin{figure}
	\centering
	\includegraphics[width=0.48\textwidth]{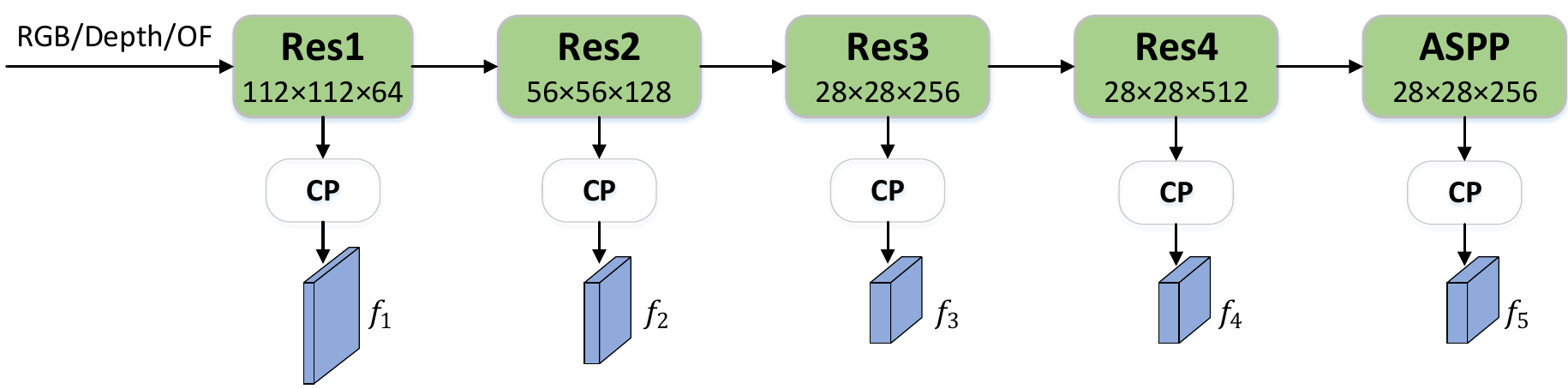}
    \vspace{-0.3cm}
	\caption{\small {Encoder of the proposed SMFNet. CP denotes channel compression.}}
    \label{fig_encoder}
    \vspace{-0.4cm}
\end{figure}

The overall framework of the proposed SMFNet is illustrated in Fig. \ref{fig_overview}, which is a trimodal network including RGB, optical flow and depth branches. Note the optical flow maps are rendered by RAFT \cite{teed2020raft}. To make a fair comparison with previous trimodal RGB-D VSOD methods \cite{lu2022depth,mou2023salient}, we apply the same encoder as them. As shown in Fig. \ref{fig_encoder}, the encoder adopts ResNet-34 \cite{he2016deep} as backbone and an Atrous Spatial Pyramid Pooling (ASPP) \cite{chen2017deeplab} module is attached to the last layer. Raw images of RGB/depth/optical flow are fed to the encoders to produce five-level features, which are denoted as $
f_{i}^{m}=\left\{ f_{i}^{m},m\in \left[ r,f,d \right] ,i=1,2,...,5 \right\} 
$. For computational convenience, we use a compression module (CP) to set the channels of all features to 64. In order to prevent excessive information loss, the CP module adopts 1$\times$1 convolution to compress the number of channels. After encoding, instead of directly fusing trimodal features, we first use PSF to conduct selective fusion of depth features $
f_{i}^{d}\left( i=1,2,...,5 \right) 
$ and optical flow features $
f_{i}^{f}\left( i=1,2,...,5 \right) 
$ to obtain a group of new features $
f_{i}^{df}\left( i=1,2,...5 \right) 
$. Then, the fused features $
f_{i}^{df}$ and the extracted RGB features $
f_{i}^{r}$ are fed to MSAM to achieve comprehensive integration of cross-modal features through perceptual interactions at multiple dimensions. Finally, the hierarchical integrated features are fed to a U-Net structure for aggregation. Details of the modules are described below.

\subsection{Pixel-level Selective Fusion Strategy}\label{sec32}

As mentioned in Sec. \ref{sec1}, effectively utilizing optical flow and depth requires considering their actual contributions. To this end, we design a pixel-level selective fusion strategy (PSF) to select and fuse the most valuable features of optical flow and depth. Fig. \ref{fig_PSF} shows the diagram of PSF, which includes two components, the generation of a spatial weight map (``SW Generation'') and a pseudo-supervisory algorithm.

After feature extraction of optical flow and depth, we obtain two groups of five-level features, i.e., $
f_{i}^{f}\left( i=1,2,...,5 \right)$ and $
f_{i}^{d}\left( i=1,2,...,5 \right)$, and then these features are fed to PSF. In order to achieve the interaction of optical flow and depth, each level of individual features are concatenated in channel dimension and processed by convolutions. To integrate features of different levels, we first use $\times$2 bilinear interpolation to enlarge the scale of high-level features, and then concatenate it to low-level features. By repeating the above steps, we can aggregate all extracted features of optical flow and depth. In the last convolutional layer, we use a $Sigmoid$ activation function to process the aggregated features to obtain a normalized weight map $SW$, which is used to weigh and sum optical flow and depth features. To match the size of hierarchical features, we resize $SW$ to get a group of spatial weight maps ${SW}_{i}\left( i=1,2,...,5 \right)$. Then we exploit ${SW}_{i}$ to conduct selective fusion of optical flow and depth, which is defined as:
\begin{equation}\label{eq1}
f_{i}^{df}={SW}_{i}\otimes f_{i}^{f}+\left( 1-{SW}_{i} \right) \otimes f_{i}^{d},
\end{equation}
where $f_{i}^{df}$ represents the selectively fused features, and $\otimes$ denotes element-wise multiplication.

However, the spatial weight maps ${SW}_{i}$ obtained without any guidance are usually hard to reflect the actual contributions of optical flow and depth. Therefore, we design a novel pseudo-supervisory algorithm to guide the learning process of ``SW Generation'', thus generating more satisfactory spatial weight maps ${SW}_{i}$. The pseudo-supervisory algorithm is illustrated in the right part of Fig. \ref{fig_PSF}.
First, in order to perceive the potential capabilities of optical flow and depth, we integrate the hierarchical features of optical flow and depth separately through upsampling and applying convolutions, and then utilize the integrated features to predict two coarse saliency maps $S_{F}$ and $S_{D}$, which are supervised by ground truth (GT). Note that to make the coarse saliency maps $S_{F}$ and $S_{D}$ reflect their own potentials, we need to pre-train optical flow stream and depth stream. The pre-training process will be detailed in Sec. \ref{sec42}. 
Next, we normalize $S_{F}$, $S_{D}$ and GT into interval $\left[ 0,1 \right]$, in which ``1'' corresponds to salient pixels whereas ``0'' corresponds to non-salient pixels.
Finally, we perform pixel-level calculation over $S_{F}$, $S_{D}$ and GT to obtain pseudo ground truth ($pGT$) as follows:

\begin{figure}
	\centering
	\includegraphics[width=0.48\textwidth]{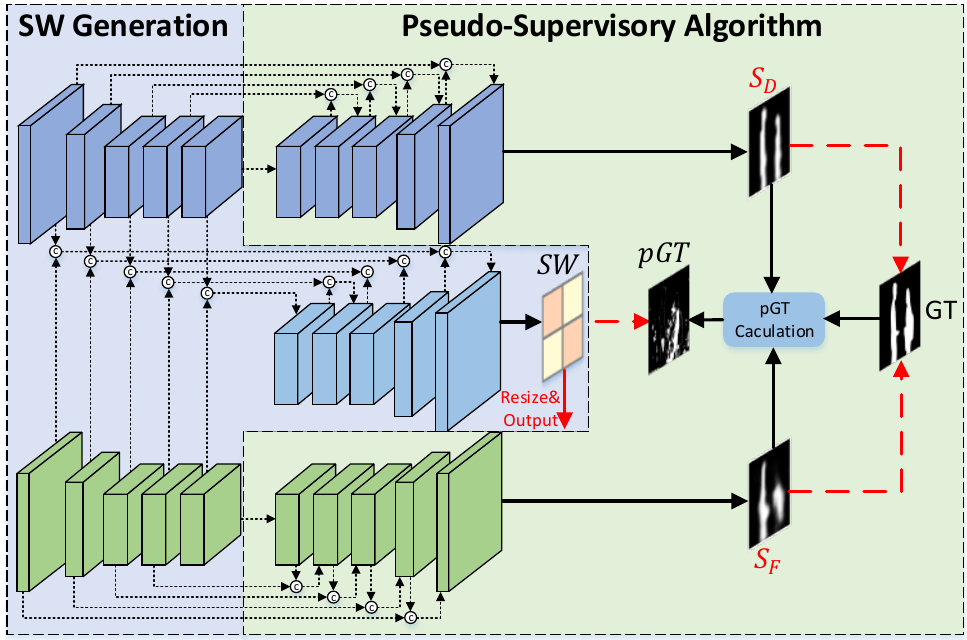}
    \vspace{-0.3cm}
	\caption{\small {Diagram of the proposed pixel-level selective fusion strategy (PSF). ``SW Generation'' means the generation of a spatial weight map $SW$. During training, PSF includes ``SW Generation'' and a pseudo-supervisory algorithm. When testing, PSF only needs the ``SW Generation'' step.}}
    \label{fig_PSF}
    \vspace{-0.4cm}
\end{figure}

(i) We evaluate the contributions of $S_{F}$ and $S_{D}$ to salient regions of GT. For each salient pixel in GT, the contributions of $S_{F}$ and $S_{D}$ to this salient pixel depend on their corresponding pixel values. That is, the larger pixel value among them means greater contribution and should be selected. Thus, the following equation is defined for pixel-wise calculation:
  \begin{equation}\label{eq2}
    pGT_s(j)=\begin{cases}
        1&		if\,\,S_F(j)>S_D(j)\,\,and\,\,GT(j)=1\\
	0&		if\,\,S_F(j) \leq S_D(j)\,\,and\,\,GT(j)=1\\
\end{cases},
  \end{equation}
where $pGT_s$ is a binary pseudo GT for supervising the positions in $SW$ that correspond to the salient regions in GT, and $j$ is the pixel index. 

\begin{figure}
	\centering
	\includegraphics[width=0.48\textwidth]{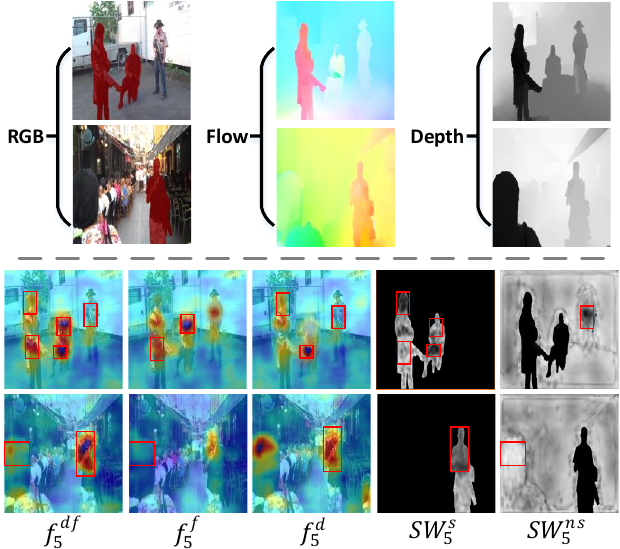}
    \vspace{-0.3cm}
	\caption{\small {Visualization of the results derived from the pixel-level selective fusion strategy. The upper part shows RGB, optical flow, and depth images, and the red-overlay regions in RGB indicate salient objects of GT. $SW_{5}^{s}$ and $SW_{5}^{ns}$ represent GT-masked parts and the counterparts in $SW_5$, respectively.}}
    \label{fig_visual}
    \vspace{-0.4cm}
\end{figure}

(ii) We evaluate the contributions of $S_{F}$ and $S_{D}$ to non-salient regions of GT. For each non-salient pixel in GT, the contributions of $S_{F}$ and $S_{D}$ to this non-salient pixel also depend on their corresponding pixel values. That is, the smaller pixel value among them means greater contribution and should be selected. Thus, the following equation is defined for pixel-wise calculation:
  \begin{equation}\label{eq3}
      pGT_{ns}(j)=\begin{cases}
	1&		if\,\,S_F(j)<S_D(j)\,\,and\,\,GT(j)=0\\
	0&		if\,\,S_F(j) \geq S_D(j)\,\,and\,\,GT(j)=0\\
\end{cases},
  \end{equation}
where $pGT_{ns}$ is also a binary pseudo GT for supervising the positions in $SW$ that correspond to the non-salient regions in GT, and $j$ is the pixel index. 
  
(iii) The final $pGT$ can be obtained by the union of $pGT_s$ and $pGT_{ns}$:
  \begin{equation}\label{eq4}
      pGT={pGT_s}\cup {pGT_{ns}},
  \end{equation}
where $pGT$ is an overall binary pseudo GT for supervising the entire spatial weight map ${SW}$. In $pGT$, a pixel value of 1 means that at this position we should trust and select optical flow information, while a pixel value of 0 means that we should select the depth counterpart at this position.

During training, we use $pGT$ to supervise the spatial weight map ${SW}$, aiming to guarantee efficacy and correctness of ${SW}$. When testing, we only need the process of ``SW Generation'' to adaptively generate $SW$, which is used to compute the weighed sum of optical flow and depth features via Eq. (\ref{eq1}), thus achieving the optimal selective fusion of optical flow and depth based on their estimated actual contributions. 

To intuitively show that PSF strategy can achieve the optimal selectively fusion of optical flow and depth based on their actual contributions, we use heatmaps to visualize $\{f_{5}^{f}$, $f_{5}^{d}$, $f_{5}^{df}\}$, and use gray-scale maps to visualize $SW_5$ in Fig. \ref{fig_visual}. To clearly see the boundaries of salient objects, we split $SW_5$ into GT-masked parts ($SW_{5}^{s}$) and the counterparts ($SW_{5}^{ns}$) using the following formulas: 
$SW_{5}^{s}=SW_5 \otimes GT$, $SW_{5}^{ns}=SW_5 \otimes (1-GT)$.
For $f_{5}^{f}$ and $f_{5}^{d}$, we use red rectangular boxes to mark some valuable regions that we hope to select. For $SW_{5}^{s}$ and $SW_{5}^{ns}$, if the pixels within the rectangular boxes are bright, it indicates selecting $f_{5}^{f}$; Otherwise, it indicates selecting $f_{5}^{d}$. We can see that the selected parts in $SW_5^{s}$ and $SW_{5}^{ns}$ match the desired valuable regions in $f_{5}^{f}$ and $f_{5}^{d}$, demonstrating that $SW_5$ can achieve optimal selection. The fusion results after Eq. (\ref{eq1}) are indicated by $f_{5}^{df}$ in Fig. \ref{fig_visual}.

\subsection{Multi-dimensional Selective Attention Module}\label{sec33}
\begin{figure}
	\centering
	\includegraphics[width=0.48\textwidth]{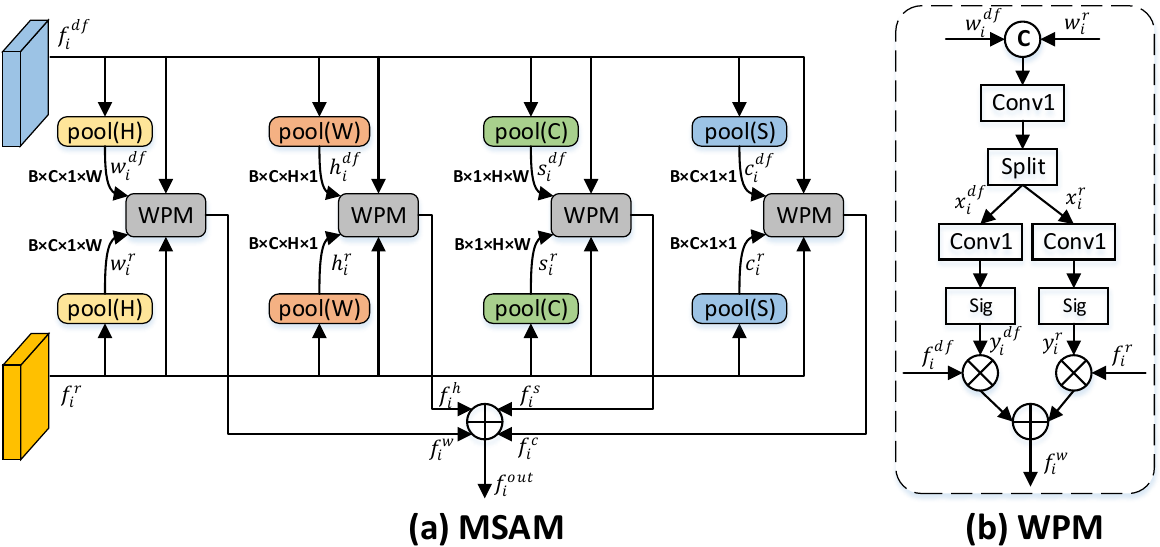}
    \vspace{-0.3cm}
	\caption{\small {(a) Overview of the proposed multi-dimensional selective attention module (MSAM). ``pool(H)'', ``pool(W)'', ``pool(C)'' and ``pool(S)'' mean average pooling operation along height, width, channel and spatial dimensions. (b) The structure of the proposed weight perception module (WPM).}}
    \label{fig_MSAM}
    \vspace{-0.4cm}
\end{figure}

After selective fusion of optical flow and depth, the next step is incorporating with RGB features to capture appearance information. However, simple fusion strategy such as concatenating and convolution can not achieve effective cross-modal interaction and generate sufficiently informative features. To fully mine the correlation across modalities, we propose a multi-dimensional selective attention module (MSAM). As shown in Fig. \ref{fig_MSAM} (a), given RGB features $f_{i}^{r}\left( i=1,2,...,5 \right)$ and also the fused features $f_{i}^{df}\left( i=1,2,...,5 \right)$ derived from PSF, MSAM conducts selective attention on width, height, spatial and channel dimensions. Specifically, we first conduct average pooling on $f_{i}^{r}$ and $f_{i}^{df}$ along height, width, channel and spatial dimensions accrodingly to compress the length of corresponding dimensions to 1 and obtain dual feature vectors. Then $f_{i}^{r}$, $f_{i}^{df}$ and the obtained feature vectors are fed to a weight perception module (WPM) to generate fused feature on a single dimension. Fig. \ref{fig_MSAM} (b) takes the selective attention at the width dimension as an example to show the rationale of WPM. To achieve feature interactions, the feature vectors $w_{i}^{df}$ and $w_{i}^{r}$ are concatenated and processed by a 1$\times$1 convolution layer to generate a mixed feature vector. We spilt this mixed feature vector into two parts, and utilize another 1$\times$1 convolution and a $Sigmoid$ activation function to process each part individually, finally generating two normalized attention vectors, which are multiplied to the original features as enhancement. Finally, the enhanced features are summed for fusion. The process of WPM can be formulated as:
\begin{align}
    & \left[ x_{i}^{df},x_{i}^{r} \right] =Split\left( Conv1\left( Cat\left( w_{i}^{df},w_{i}^{r} \right) \right) \right),\\
    & y_{i}^{df} =Sig\left( Conv1\left( x_{i}^{df} \right) \right), 
    y_{i}^{r} =Sig\left( Conv1\left( x_{i}^{r} \right) \right),\\
    & f_{i}^{w}=y_{i}^{df}\otimes f_{i}^{df}+y_{i}^{r}\otimes f_{i}^{r},
\end{align}
where $x_{i}^{df}$ and $x_{i}^{r}$ represent weight vectors, $Cat\left( \cdot \right)$ denotes a channel concatenation operation, $Conv1\left( \cdot \right)$ denotes a 1$\times$1 convolution operation, $Split\left( \cdot \right)$ denotes a split operation to yield two vectors, $Sig\left( \cdot \right)$ denotes a $Sigmoid$ activation function, and $f_{i}^{w}$ denotes the fused feature on width dimension. 

Likewise, the selective attentions on other dimensions are similarly conducted. The final fusion is formulated as:
\begin{equation}
    f_{i}^{out}=f_{i}^{w}+f_{i}^{h}+f_{i}^{s}+f_{i}^{c},
\end{equation}
where $f_{i}^{out}$ means the final fused features, $f_{i}^{h}$, $f_{i}^{s}$, and $f_{i}^{c}$ denote the fused features on height, spatial, and channel dimensions, respectively.

\subsection{Loss Function}\label{sec34}
Inspired by \cite{lu2022depth}, we adopt a combination of widely used binary cross entropy (BCE) loss and intersection-over-union (IoU) \cite{rahman2016optimizing} loss for training SMFNet, which is formulated as:
\begin{equation}
    \mathcal{L} =\mathcal{L} _{bce}+\mathcal{L} _{iou}.
\end{equation}

The total loss function consists of two parts. For the first part, as mentioned in Sec. \ref{sec32}, the coarse saliency maps $S_D$ and $S_F$ are supervised by GT, and the spatial weight map $SW$ is supervised by $pGT$. Let $\mathcal{L} _{PSF}$ be the loss of the first part, which can be formulated as:
\begin{equation}
    \mathcal{L} _{PSF}=\mathcal{L} \left( S_D,GT \right) +\mathcal{L} \left( S_F,GT \right) +\mathcal{L} \left( SW,pGT \right).
\end{equation}
For the second part, as shown in Fig. \ref{fig_overview}, the decoder predicts five saliency maps $S_{i}\left( i=1,2,...,5 \right)$. Let $\mathcal{L} _{decoder}$ be the loss of the second part, which can be formulated as:
\begin{equation}
    \mathcal{L} _{decoder}=\sum_{i=1}^5{\left( 1/2^{i-1} \right) \mathcal{L} \left( S_i,GT \right)},
\end{equation}
where ${1/2^{i-1}|_{i=1,2,...,5}}$ is the weight that we set to balance each level of loss. Note that during inference, we only take $S_1$ as the final saliency prediction.

Finally, the total loss function $\mathcal{L} _{total}$ is defined as the sum of $\mathcal{L} _{PSF}$ and $\mathcal{L} _{decoder}$:
\begin{equation}
\begin{aligned}
    \mathcal{L} _{total}=\mathcal{L} _{PSF} +\mathcal{L} _{decoder}.   
\end{aligned}
\end{equation}

\begin{table*}[t!]
    \renewcommand{\arraystretch}{0.9}
    \caption{\small Quantitative results of state-of-the-art RGB-D VSOD, VSOD and RGB-D methods on the test set of two public RGB-D video datasets. The best three results are represented in \textcolor{red}{red}, \textcolor{darkgreen}{green}, and \textcolor{blue}{blue}. $\uparrow$/$\downarrow$ indicates that the larger/smaller value is better. Notation $^\dagger$ indicates those results by fine-tuning on the joint training set of RDVS and DVisal. Symbol {$\divideontimes$} means the RGB-D VSOD field, and `**' means that the results are not available.
    }\label{tab_sota}
    \vspace{-8pt}
  \centering
    \footnotesize
    \setlength{\tabcolsep}{1.2mm}
    
    \begin{tabular}{c|c|ccc|ccc|c|ccc|ccc}
    \hline
    &\multirow{2}{*}{Methods}&\multicolumn{3}{c|}{RDVS \cite{mou2023salient}}&
    \multicolumn{3}{c|}{DVisal \cite{li2024dvsod}}
    &\multirow{2}{*}{Methods}&\multicolumn{3}{c|}{RDVS \cite{mou2023salient}}&
    \multicolumn{3}{c}{DVisal \cite{li2024dvsod}}\cr\cline{3-8} \cline{10-15}
    &&$S_\alpha\uparrow$&$F_{\beta}^{\textrm{max}}\uparrow$&$M\downarrow$&$S_\alpha\uparrow$&$F_{\beta}^{\textrm{max}}\uparrow$&$M\downarrow$&&$S_\alpha\uparrow$&$F_{\beta}^{\textrm{max}}\uparrow$&$M\downarrow$&$S_\alpha\uparrow$&$F_{\beta}^{\textrm{max}}\uparrow$&$M\downarrow$\cr
    \hline

    \multirow{8}{*}{\begin{sideways}{RGB-D SOD}\end{sideways}}

    &BBSNet \cite{fan2020bbs}&0.732&0.549&0.056&0.715&0.609&0.118
    &BBSNet$^\dagger$ \cite{fan2020bbs}&0.745&0.605&0.055&0.775&0.716&0.082 \cr

    &UC-Net \cite{zhang2020uc}&0.709&0.531&0.062&0.669&0.579&0.129
    &UC-Net$^\dagger$ \cite{zhang2020uc}&0.749&0.613&0.054&0.741&0.702&0.085 \cr

    &JL-DCF \cite{fu2020jl}&0.725&0.559&0.067&0.658&0.560&0.128
    &JL-DCF$^\dagger$ \cite{fu2020jl}&0.762&0.633&0.054&0.739&0.705&0.080 \cr

    &TriTransNet \cite{liu2021tritransnet}&0.728&0.559&0.060&0.633&0.527&0.133
    &TriTransNet$^\dagger$ \cite{liu2021tritransnet}&0.720&0.558&0.069&0.700&0.645&0.092 \cr

    &SPNet \cite{zhou2021specificity}&0.736&0.570&0.063&0.698&0.608&0.113
    &SPNet$^\dagger$ \cite{zhou2021specificity}&0.748&0.611&0.054&0.790&0.741&0.071 \cr

    &CIRNet \cite{cong2022cir}&0.736&0.582&0.060&0.663&0.595&0.108
    &CIRNet$^\dagger$ \cite{cong2022cir}&0.745&0.629&0.057&0.784&0.729&0.077 \cr

    &HRTransNet \cite{tang2022hrtransnet}&0.718&0.546&0.057&0.678&0.591&0.114
    &HRTransNet$^\dagger$ \cite{tang2022hrtransnet}&0.739&0.588&0.059&0.722&0.679&0.089 \cr

    &RD3D \cite{chen20223}&0.764&0.607&0.056&0.703&0.609&0.118
    &RD3D$^\dagger$ \cite{chen20223}&0.737&0.570&0.064&0.771&0.729&0.100 \cr

    &PICRNet \cite{cong2023point}&0.717&0.552&0.070&0.684&0.597&0.112
    &PICRNet$^\dagger$ \cite{cong2023point}&0.797&0.698&0.047&0.780&0.723&0.081 \cr

    &CPNet \cite{hu2024cross}&0.749&0.613&0.053&0.701&0.641&0.094
    &CPNet$^\dagger$ \cite{hu2024cross}&0.792&0.671&0.048&\textcolor{darkgreen}{0.835}&\textcolor{blue}{0.799}&\textcolor{darkgreen}{0.050} \cr
    \hline

     \multirow{8}{*}{\begin{sideways}{VSOD}\end{sideways}}
    
    &MGAN \cite{li2019motion}&0.826&0.736&0.043&\textcolor{darkgreen}{0.745}&\textcolor{darkgreen}{0.712}&\textcolor{darkgreen}{0.082}
    &MGAN$^\dagger$ \cite{li2019motion}&0.827&0.739&0.043&0.783&0.731&0.076 \cr

    &STVS \cite{chen2021exploring}&0.766&0.648&0.049&0.714&0.650&0.099
    &STVS$^\dagger$ \cite{chen2021exploring}&0.754&0.634&0.057&0.697&0.639&0.096 \cr

    &WSVSOD \cite{zhao2021weakly}&0.702&0.563&0.73&0.610&0.509&0.148
    &WSVSOD$^\dagger$ \cite{zhao2021weakly}&**&**&**&**&**&** \cr

    &FSNet \cite{ji2021full}&0.824&0.745&0.046&0.697&0.653&0.100
    &FSNet$^\dagger$ \cite{ji2021full}&0.816&0.739&0.048&0.722&0.676&0.091 \cr

    &DCFNet \cite{zhang2021dynamic}&0.768&0.647&0.049&0.720&0.674&0.092
    &DCFNet$^\dagger$ \cite{zhang2021dynamic}&0.790&0.662&0.048&0.743&0.701&0.086 \cr

    &UGPL \cite{piao2022semi}&0.772&0.669&0.049&0.709&0.598&0.119
    &UGPL$^\dagger$ \cite{piao2022semi}&0.797&0.692&0.049&0.752&0.718&0.080 \cr
    \hline
    
     \multirow{4}{*}{$\divideontimes$}

    &DVSOD \cite{li2024dvsod}&0.698&0.508&0.066&0.729&0.669&0.113
    &DVSOD$^\dagger$ \cite{li2024dvsod}&0.717&0.544&0.057&0.732&0.671&0.108 \cr

    &ATFNet \cite{lin2024vidsod}&0.712&0.584&0.062&0.703&0.608&0.115
    &ATFNet$^\dagger$ \cite{lin2024vidsod}&0.749&0.595&0.054&0.727&0.665&0.112 \cr

    &DCTNet \cite{lu2022depth}&\textcolor{blue}{0.846}&\textcolor{blue}{0.780}&\textcolor{darkgreen}{0.033}&0.727&0.676&\textcolor{blue}{0.084}
    &DCTNet$^\dagger$ \cite{lu2022depth}&\textcolor{blue}{0.849}&\textcolor{blue}{0.785}&\textcolor{darkgreen}{0.033}&0.822&0.796&0.054 \cr

    &DCTNet+ \cite{mou2023salient}&\textcolor{darkgreen}{0.861}&\textcolor{darkgreen}{0.803}&\textcolor{blue}{0.036}&\textcolor{blue}{0.738}&\textcolor{blue}{0.696}&0.091
    &DCTNet+$^{\dagger}$ \cite{mou2023salient}&\textcolor{darkgreen}{0.866}&\textcolor{darkgreen}{0.812}&\textcolor{blue}{0.035}&\textcolor{blue}{0.833}&\textcolor{darkgreen}{0.823}&\textcolor{blue}{0.052} \cr

    &SMFNet&\textcolor{red}{0.874}&\textcolor{red}{0.823}&\textcolor{red}{0.028}&\textcolor{red}{0.854}&\textcolor{red}{0.851}&\textcolor{red}{0.038}
    &SMFNet&\textcolor{red}{0.874}&\textcolor{red}{0.823}&\textcolor{red}{0.028}&\textcolor{red}{0.854}&\textcolor{red}{0.851}&\textcolor{red}{0.038} \cr

    \hline
    \end{tabular}
\vspace{-0.2cm}
\end{table*}

\section{EXPERIMENTS and results}\label{Experiments_Results}
\subsection{Datasets and Metrics}\label{sec41}
Since the RGB-D video datasets RDVS \cite{mou2023salient} and DVisal \cite{li2024dvsod} are recently proposed, no models have been experimented on both datasets. As a result, the proposed SMFNet is the first model to be experimented on both RDVS and DVisal. To achieve more comprehensive evaluation of SMFNet, we not only compare with existing RGB-D VSOD methods, but also select some representative RGB-D and VSOD methods for comparisons. Specifically, for evaluating SMFNet on RDVS (4,030 frames) and DVisal (7,117 frames), we merge the training set of RDVS (2,176 frames) and the training set of DVisal (3,551 frames) for training. The remaining samples are used for testing. Note that since the last frame of each sequence lacks the corresponding optical flow, we do not test such frames. For more comparisons, we use DPT \cite{ranftl2021vision} to generate synthetic depth maps for five VSOD benchmark datasets, namely, DAVIS \cite{perazzi2016benchmark}, DAVSOD \cite{fan2019shifting}, FBMS \cite{ochs2013segmentation}, SegTrack-V2 \cite{li2013video} and VOS \cite{li2017benchmark}, and conduct experiments on them. Following \cite{mou2023salient}, we choose 7,683 frames from DAVIS, DAVSOD and FBMS as our training sets, and 9,502 frames from above five VSOD datasets for testing. Three widely used saliency metrics are adopted for evaluation, including S-measure ($S_{\alpha}$) \cite{fan2017structure}, maximum F-measure ($F_{\beta}^{max}$) \cite{borji2015salient,achanta2009frequency} and MAE ($M$) \cite{borji2015salient,perazzi2012saliency}. Higher $S_{\alpha}$, $F_{\beta}^{max}$, and lower $M$ indicate better performance. 

\begin{figure*}
	\centering
	\includegraphics[width=0.98\textwidth]{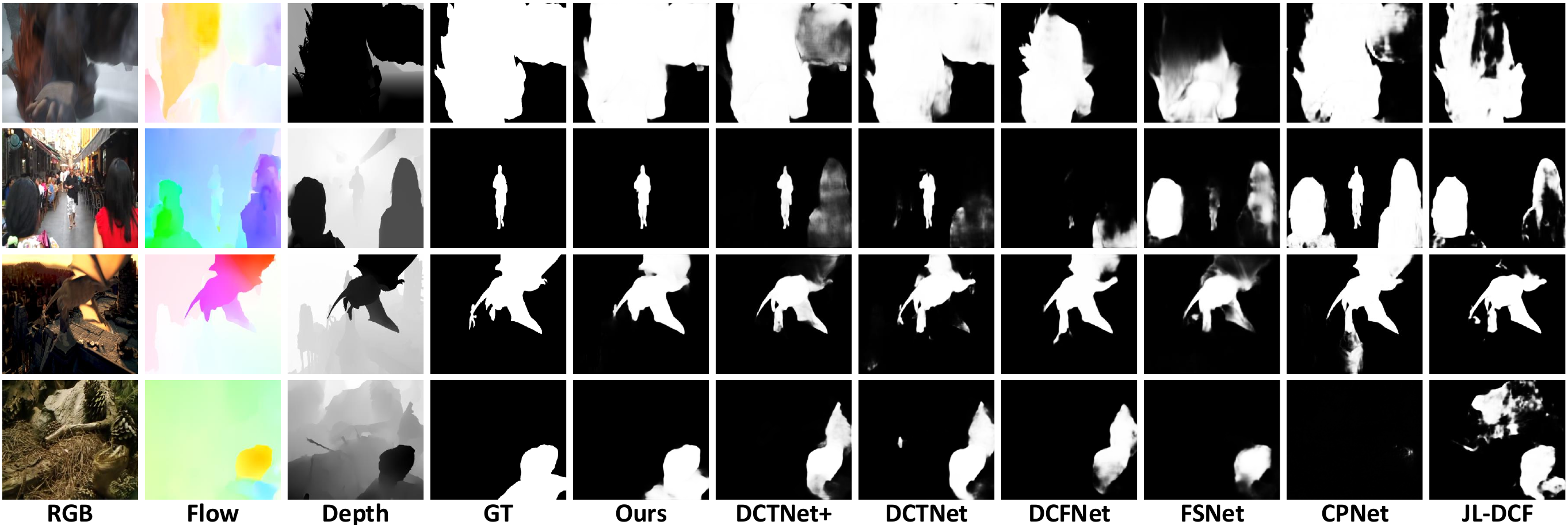}
    \vspace{-0.3cm}
	\caption{\footnotesize{ Visual comparison with 6 SOTA models (including DCTNet+ \cite{mou2023salient}, DCTNet \cite{lu2022depth}, DCFNet \cite{zhang2021dynamic}, FSNet \cite{ji2021full}, CPNet \cite{hu2024cross}, JL-DCF \cite{fu2020jl}). It can be seen that our SMFNet performs the best in many challenging scenarios.}} \label{fig_quality}
    \vspace{-0.4cm}
\end{figure*}

\subsection{Implementation Details}\label{sec42}
Our SMFNet is implemented in PyTorch, and was trained on an NVIDIA 4090 GPU. As mentioned in Sec. \ref{sec32}, to make the coarse saliency maps $S_D$ and $S_F$ reflect their own potentials, we first pre-train the depth stream of PSF together with the depth encoder of SMFNet by feeding depth maps. This similar procedure is also applied to the optical flow counterpart. To ensure consistency in the training process, the RGB encoder of SMFNet is also pre-trained by feeding RGB images. We adopt a U-Net structure to integrate hierarchical features of RGB and predict coarse saliency maps, which is supervised by GT. Next, we fine-tune the entire SMFNet model on the whole training samples. During training, the initial learning rates of backbones and other parts are set to 1e-5 and 1e-4, respectively. The SGD optimizer is adopted under the batch size 8. All input images are uniformly resized to 448$\times$448 for training and testing, and are also augmented using various strategies like random flipping, random cropping and random rotating during training. The model converges after 70 training epochs.

\subsection{Comparisons with State-of-the-Arts on RDVS and DVisal}\label{sec43}

To validate the effectiveness of the proposed SMFNet, we
quantitatively compare it with four existing RGB-D VSOD methods (DVSOD \cite{li2024dvsod}, ATFNet \cite{lin2024vidsod}, DCTNet \cite{lu2022depth}, DCTNet+ \cite{mou2023salient}), six SOTA VSOD methods (MGAN \cite{li2019motion}, UGPL \cite{piao2022semi}, STVS \cite{chen2021exploring}, WSVSOD \cite{zhao2021weakly}, FSNet \cite{ji2021full}, DCFNet \cite{zhang2021dynamic}) and 10 SOTA RGB-D SOD methods (CPNet \cite{hu2024cross}, PICRNet \cite{cong2023point}, RD3D \cite{chen20223}, HRTransNet \cite{tang2022hrtransnet}, CIRNet \cite{cong2022cir}, \cite{zhou2021specificity}, TriTransNet \cite{liu2021tritransnet}, JL-DCF \cite{fu2020jl}, UC-Net \cite{zhang2020uc}, BBSNet \cite{fan2020bbs}). For fairer and more comprehensive comparison, we evaluate not only the original models, but also the fine-tuned models re-trained on the joint training set of RDVS and DVisal. Note that the above chosen methods for comparison are all open-source, in order to make the experiments feasible.

\begin{table*}[t!]
    \renewcommand{\arraystretch}{0.9}
    \caption{\small Quantitative comparison with state-of-the-art VSOD methods on 5 benchmark datasets. The best three results are represented in \textcolor{red}{red}, \textcolor{darkgreen}{green}, and \textcolor{blue}{blue}. $\uparrow$/$\downarrow$ indicates that the larger/smaller value is better. Symbol `**' means that results are not available.
    }\label{tab_sota_vsod}
    \vspace{-8pt}
  \centering
    \footnotesize
    \setlength{\tabcolsep}{1.2mm}
    
    \begin{tabular}{c|ccc|ccc|ccc|ccc|ccc}
    \hline
    \multirow{2}{*}{Methods}&\multicolumn{3}{c|}{DAVIS\cite{perazzi2016benchmark}}&
    \multicolumn{3}{c|}{DAVSOD\cite{fan2019shifting}}&\multicolumn{3}{c|}{FBMS\cite{ochs2013segmentation}}&\multicolumn{3}{c|}{SegV2\cite{li2013video}}&\multicolumn{3}{c}{VOS\cite{li2017benchmark}}\cr\cline{2-16}
    &$S_\alpha\uparrow$&$F_{\beta}^{\textrm{max}}\uparrow$&$M\downarrow$&$S_\alpha\uparrow$&$F_{\beta}^{\textrm{max}}\uparrow$&$M\downarrow$&$S_\alpha\uparrow$&$F_{\beta}^{\textrm{max}}\uparrow$&$M\downarrow$&$S_\alpha\uparrow$&$F_{\beta}^{\textrm{max}}\uparrow$&$M\downarrow$&$S_\alpha\uparrow$&$F_{\beta}^{\textrm{max}}\uparrow$&$M\downarrow$\cr
    \hline
    
    FGRNE\cite{li2018flow}&0.838&0.783&0.043&0.701&0.589&0.095&0.809&0.767&0.088&0.770&0.694&0.035&0.715&0.669&0.097 \cr
    PDBM\cite{song2018pyramid}&0.882&0.855&0.028&0.706&0.591&0.114&0.851&0.821&0.064&0.864&0.808&0.024&0.818&0.742&0.078 \cr
    SSAV\cite{fan2019shifting}&0.893&0.861&0.028&0.755&0.659&0.084&0.879&0.865&0.040&0.851&0.798&0.023&0.786&0.704&0.091 \cr
    MGAN\cite{li2019motion}&0.913&0.893&0.022&0.757&0.663&0.079&\textcolor{blue}{0.912}&0.909&0.026&0.895&0.840&0.024&0.807&0.743&0.069 \cr
    PCSA\cite{gu2020pyramid}&0.902&0.880&0.022&0.741&0.656&0.086&0.868&0.837&0.040&0.866&0.811&0.024&0.828&0.747&0.065 \cr
    TENet\cite{ren2020tenet}&0.905&0.881&0.017&0.779&0.697&0.070&\textcolor{darkgreen}{0.916}&\textcolor{blue}{0.915}&\textcolor{darkgreen}{0.024}&**&**&**&**&**&** \cr
    STVS\cite{chen2021exploring}&0.892&0.865&0.023&0.746&0.651&0.086&0.872&0.856&0.038&0.891&0.860&0.017&\textcolor{blue}{0.850}&0.791&0.058 \cr
    WSVSOD\cite{zhao2021weakly}&0.846&0.793&0.038&0.694&0.593&0.115&0.803&0.792&0.073&0.819&0.762&0.033&0.765&0.702&0.089 \cr
    FSNet\cite{ji2021full}&0.920&0.907&0.020&0.773&0.685&0.072&0.890&0.888&0.041&0.870&0.806&0.025&0.703&0.659&0.108 \cr
    DCFNet\cite{zhang2021dynamic}&0.914&0.900&0.016&0.741&0.660&0.074&0.873&0.840&0.039&0.883&0.839&0.015&0.845&0.791&0.061 \cr

    MGTNet\cite{min2022mutual}&\textcolor{blue}{0.925}&\textcolor{blue}{0.918}&0.015&0.796&0.721&0.064&0.901&0.890&0.033&0.893&0.849&\textcolor{blue}{0.014}&0.835&0.766&0.062 \cr
    
    UGPL \cite{piao2022semi}&0.910&0.895&0.020&0.749&0.658&0.074&0.900&0.892&0.027&0.860&0.803&0.025&0.764&0.706&0.078 \cr

    CoSTFormer \cite{liu2023learning}&0.921&0.903&\textcolor{blue}{0.014}&\textcolor{blue}{0.806}&\textcolor{blue}{0.731}&\textcolor{blue}{0.061}&0.889&0.885&0.036&\textcolor{blue}{0.904}&\textcolor{blue}{0.870}&0.016&0.812&0.748&0.081 \cr

    ATFNet \cite{lin2024vidsod}&0.901&0.886&{0.020}&{0.747}&{0.660}&{0.075}&0.863&0.825&0.046&{0.842}&{0.794}&0.028&0.802&0.733&0.095 \cr
    
    DCTNet \cite{lu2022depth}&0.922&0.912&0.015&0.797&0.728&\textcolor{blue}{0.061}&0.911&0.913&\textcolor{blue}{0.025}&0.889&0.840&0.019&0.846&\textcolor{blue}{0.793}&\textcolor{darkgreen}{0.051} \cr
    
    DCTNet+ \cite{mou2023salient} &\textcolor{darkgreen}{0.930}&\textcolor{darkgreen}{0.922}&\textcolor{darkgreen}{0.012}&\textcolor{darkgreen}{0.818}&\textcolor{darkgreen}{0.754}&\textcolor{darkgreen}{0.055}&\textcolor{darkgreen}{0.916}&\textcolor{darkgreen}{0.918}&0.026&\textcolor{red}{0.931}&\textcolor{darkgreen}{0.917}&\textcolor{red}{0.010}&\textcolor{darkgreen}{0.858}&\textcolor{darkgreen}{0.802}&\textcolor{blue}{0.056} \cr
    
    SMFNet &\textcolor{red}{0.937}&\textcolor{red}{0.932}&\textcolor{red}{0.011}&\textcolor{red}{0.833}&\textcolor{red}{0.781}&\textcolor{red}{0.045}&\textcolor{red}{0.923}&\textcolor{red}{0.934}&\textcolor{red}{0.022}&\textcolor{darkgreen}{0.928}&\textcolor{red}{0.918}&\textcolor{darkgreen}{0.011}&\textcolor{red}{0.860}&\textcolor{red}{0.821}&\textcolor{red}{0.046} \cr
    \hline
    \end{tabular}
\vspace{-0.4cm}
\end{table*}

\textit{1) Quantitative Comparison:} Table \ref{tab_sota} shows the results of original models and fine-tuned models evaluated on the test sets of RDVS and DVisal. Firstly, it can be seen that the proposed SMFNet outperforms all existing RGB-D VSOD methods. Moreover, compared with SOTA RGB-D VSOD methods (i.e., original DCTNet+ \cite{mou2023salient}), SMFNet achieves significant improvement on all metrics. Specifically, on RDVS, the percentage gain of SMFNet reaches 1.3$\%$ for $S_\alpha$, 2.0$\%$ for $F_{\beta}^{\textrm{max}}$ and 0.8$\%$ for $M$. On DVisal, the percentage gain of SMFNet reaches 11.6$\%$ for $S_\alpha$, 15.5$\%$ for $F_{\beta}^{\textrm{max}}$ and 5.3$\%$ for $M$. We can see SMFNet has a huge improvement on DVisal compared with original DCTNet+. The reason is that the quality of depth maps in DVisal is generally low, and the original model of DCTNet+ has not been trained on low-quality depth maps, resulting in weak performance when testing directly on DVisal. When re-trained on the joint training set of RDVS and DVisal, the performance of DCTNet+ improves on both datasets, especially on DVisal. In addition, the overall performance of almost all models will be greatly improved after being re-trained, which indicates that the joint training set can enhance the model's robustness. Nevertheless, our SMFNet still outperforms all fine-tuned models.

\textit{2) Qualitative Comparison:} Fig. \ref{fig_quality} shows visual comparison results of our SMFNet and other six SOTA models on challenging scenarios, including low-quality optical flow or depth maps ($1^{st}$ and $2^{nd}$ rows), and complex and low contrast background ($3^{rd}$ and $4^{th}$ rows). From these results, we can see that our SMFNet predicts most accurately on salient objects, fully demonstrating the robustness and effectiveness of SMFNet against various chaotic information.

\subsection{Comparisons with State-of-the-Arts on VSOD Benchmarks}\label{sec44}

Since our proposed SMFNet is the first model to be evaluated on RDVS and DVisal, the experimental results in Table \ref{tab_sota} may need extra support to prove the effectiveness of SMFNet. To this end, we compare SMFNet with 15 deep learning-based methods on conventional VSOD benchmarks. However, VSOD benchmarks do not have available realistic depth maps, so we follow the previous literature \cite{lu2022depth,mou2023salient} to generate a synthetic depth map for each video frame. Note that all the training details are kept unchanged as those in Sec. \ref{sec42}. Quantitative results on five VSOD benchmark datasets are shown in Table \ref{tab_sota_vsod}. We can see encouraging improvement of SMFNet over most VSOD methods. Specifically, compared with the latest DCTNet+ \cite{mou2023salient}, SMFNet gains 1.5$\%$ on $S_\alpha$, 2.7$\%$ on $F_{\beta}^{\textrm{max}}$, and 1$\%$ on $M$ over the largest VSOD dataset, i.e., DAVSOD \cite{fan2019shifting}. On the other four benchmarks, SMFNet also outperforms almost all VSOD methods, fully demonstrating the superiority of SMFNet.

\subsection{Ablation Study}\label{sec45}
In this section, we conduct a series of ablation experiments on RDVS and DVisal datasets to verify the effectiveness of different components in the proposed SMFNet.

\textit{1) Effectiveness of the modules:} To validate the effectiveness of the proposed modules in SMFNet and show their performance gains, we start from a baseline model and gradually extend it with different modules, including PSF and MSAM. As shown in Table \ref{Ablation1}, four component settings are evaluated. The first setting only includes baseline, which is implemented by replacing PSF and MSAM in SMFNet with concatenation and convolution operation. The second setting adds PSF upon the baseline, improving the model performance significantly. The third setting only replaces PSF in SMFNet with concatenation and convolution operation, which also outperforms baseline. The last setting is our proposed SMFNet, consisting of baseline, PSF and MSAM, which achieves the best performance on all metrics and outperforms baseline a lot.

\textit{2) Effectiveness of each component in PSF:} Table \ref{Ablation1} demonstrates that PSF contributes to the superior performance of SMFNet. To reveal the contribution of each component in PSF, we first evaluate the proposed SMFNet without PSF, i.e., baseline (without PSF) in Table \ref{Ablation2}. Then we generate a spatial weight map $SW$ shown in Fig. \ref{fig_PSF} to perform a weighted sum of optical flow and depth (+$SW$). However, the model performance is not significantly improved because $SW$ can not effectively select the most valuable optical flow and depth features without the guidance of $pGT$. Model ``+ $SW$ + $pGT$'' adds our proposed pseudo-supervisory algorithm upon model ``+ $SW$'' and forms the complete PSF. The experimental results show that model ``+ $SW$ + $pGT$'' can achieve better performance than model ``+ $SW$'', which means that the supervision of $pGT$ is effective, and the generated spatial weight map $SW$ is beneficial.

\begin{table}[t!]
	\centering
	\footnotesize
	\renewcommand\arraystretch{0.9}
	\caption{Ablation study of each module in SMFNet. The best results are shown in bold}
    \vspace{-8pt}
	\setlength{\tabcolsep}{1mm}{
		\begin{tabular}{ccc|ccc|ccc}
			\toprule[1.5pt]
			\multicolumn{3}{c|}{Component Setting}    & \multicolumn{3}{c}{RDVS\cite{mou2023salient}} & \multicolumn{3}{|c}{DVsial\cite{li2024dvsod}}\\
			\cmidrule{1-9}
			{baseline}& {PSF} & {MSAM}&
			\multicolumn{1}{c}{$S_\alpha\uparrow$} & \multicolumn{1}{c}{$F_{\beta}^{\textrm{max}}\uparrow$} & \multicolumn{1}{c}{$M\downarrow$} & \multicolumn{1}{|c}{$S_\alpha\uparrow$} & \multicolumn{1}{c}{$F_{\beta}^{\textrm{max}}\uparrow$} & \multicolumn{1}{c}{$M\downarrow$}
			\\
			\midrule
			\ding{52}&       &       &        0.861 & 0.810 & 0.033 & 0.837  & 0.839&  0.044  \\
			
			\ding{52}&   \ding{52}    &         &0.869  & 0.817 & 0.030 & 0.849 & 0.850 & 0.039  \\
			
			\ding{52}&   &    \ding{52}      & 0.866  &  0.815 & 0.031 & 0.848 & 0.846 &  0.040  \\
			
			\midrule
			\ding{52}&   \ding{52}    &     \ding{52} & \textbf{0.874} & \textbf{0.823} & \textbf{0.028} & \textbf{0.854} & \textbf{0.851} & \textbf{0.038} \\
			\toprule[1.5pt]
			\\
	\end{tabular}}%
	\label{Ablation1}%
 `  \vspace{-0.6cm}
\end{table}%

\begin{table}[t!]
	\centering
	\footnotesize
	\renewcommand\arraystretch{0.9}
	\caption{Ablation study of each component in PSF. The best results are shown in bold}
    \vspace{-8pt}
	\setlength{\tabcolsep}{1mm}{
		\begin{tabular}{l|ccc|ccc}
			\toprule[1.5pt]
            \multirow{2}[3]{*}{Model}& \multicolumn{3}{c}{RDVS\cite{mou2023salient}} & \multicolumn{3}{|c}{DVsial\cite{li2024dvsod}}\\
			\cmidrule{2-7}
			&$S_\alpha\uparrow$ & $F_{\beta}^{\textrm{max}}\uparrow$ & $M\downarrow$ & $S_\alpha\uparrow$ & $F_{\beta}^{\textrm{max}}\uparrow$ & $M\downarrow$
			\\
			\midrule
			1. baseline (without PSF)& 0.866 & 0.815 & 0.031 & 0.848  & 0.846&  0.040 \\
			
			2. +$SW$ & 0.871  & 0.818 & 0.032 & 0.846 & 0.843 &  0.039  \\
			
			3. +$SW$+$pGT$ (with PSF)&\textbf{0.874} & \textbf{0.823} & \textbf{0.028} & \textbf{0.854} & \textbf{0.851} & \textbf{0.038} \\
			\toprule[1.5pt]
	\end{tabular}}%
	\label{Ablation2}%
    \vspace{-0.5cm}
\end{table}%

\begin{table}[t!]
	\centering
	\footnotesize
	\renewcommand\arraystretch{0.9}
	\caption{Ablation study of cross-modal fusion in PSF. The best results are shown in bold}
    \vspace{-8pt}
	\setlength{\tabcolsep}{0.8mm}{
		\begin{tabular}{l|ccc|ccc}
			\toprule[1.5pt]
            \multirow{2}[3]{*}{Model}& \multicolumn{3}{c}{RDVS\cite{mou2023salient}} & \multicolumn{3}{|c}{DVsial\cite{li2024dvsod}}\\
			\cmidrule{2-7}
			&$S_\alpha\uparrow$ & $F_{\beta}^{\textrm{max}}\uparrow$ & $M\downarrow$ & $S_\alpha\uparrow$ & $F_{\beta}^{\textrm{max}}\uparrow$ & $M\downarrow$
			\\
			\midrule
			1. PSF (RGB, Optical flow)& 0.868 & 0.814 & 0.032 & 0.845  & 0.842&  0.040  \\
			
			2. PSF (RGB, Depth) & 0.870  & 0.817 & 0.031 & 0.848 & 0.849 &   \textbf{0.038}  \\
			
			3. PSF (Optical flow, Depth) &\textbf{0.874} & \textbf{0.823} & \textbf{0.028} & \textbf{0.854} & \textbf{0.851} & \textbf{0.038} \\
			\toprule[1.5pt]
	\end{tabular}}%
	\label{Ablation3}%
    \vspace{-0.4cm}
\end{table}%

\textit{3) Effectiveness of the fusion of optical flow and depth:} To verify that the fusion of optical flow and depth in PSF can unleash the power of motion and depth, thus improving the model's performance, we try two other cross-modal fusions in PSF, namely ``PSF (RGB, Optical flow)'' and ``PSF (RGB, Depth)''. Note that when conducting ``PSF (RGB, Optical flow)''/``PSF (RGB, Depth)'', depth/optical flow is used to replace RGB input of MSAM. Table \ref{Ablation3} shows that the experimental results of different cross-modal fusions in PSF. We can see that compared with ``PSF (RGB, Optical flow)'' and ``PSF (RGB, Depth)'', our method, i.e., PSF (Optical flow, Depth), has a good improvement on all metrics. The results demonstrate that the fusion of optical flow and depth can provide more useful features, helping to detect salient objects. Note that ``PSF (RGB, Depth)'' also outperforms ``PSF (RGB, Optical flow)'', especially on DVisal dataset. The reason is that depth contain more noises than optical flow, especially in DVisal, and PSF can suppress most noises during the cross-modal fusion process, thus helping to improve the final prediction.

\textit{4) Effectiveness of each component in MSAM:} To investigate the effectiveness of each component in MSAM, we conduct a series of ablation experiments and show their results in Table \ref{Ablation4}. We replace MSAM with concatenation and convolution as our baseline, and gradually extend it with selective attention on width (W), height (H), spatial (S) and channel (C) dimensions. As shown in Table \ref{Ablation4}, every extended branch contributes to the superior performance of SMFNet to some extent. The reason is that multi-dimensional interactions can enhance feature fusion between different modalities and generate richer and more refined features. Therefore, the results validate the above effectiveness of each component in MSAM.

\textit{5) Comparisons of MSAM to other fusion modules:} To verify the effectiveness of the entire MSAM, we compare it with four different fusion modules: i.e., muti-modal feature Aggregation (MFA) proposed in SPNet \cite{zhou2021specificity}, motion guided attention (MGA) proposed in MGAN \cite{li2019motion}, refinement fusion module (RFM) proposed in DCTNet+ \cite{mou2023salient}, and cross-modal attention fusion module (CAM) proposed in CPNet \cite{hu2024cross}. More specifically, we replace MSAM with these modules respectively in our SMFNet and keep other components unchanged. Note that RFM has three input branches (RGB, optical flow and depth), so we remove one branch (depth) in RFM to match our SMFNet. The experimental results are shown in Table \ref{Ablation5}. We can see MSAM achieves the best performance compared with the remaining fusion modules, which proves MSAM's powerful ability to mine cross-modal information and enhance feature representation.

\begin{table}[t!]
	\centering
	\footnotesize
	\renewcommand\arraystretch{0.9}
	\caption{Ablation study of each component in MSAM. The best results are shown in bold}\vspace{-8pt}
	\setlength{\tabcolsep}{0.8mm}{
		\begin{tabular}{l|ccc|ccc}
			\toprule[1.5pt]
            \multirow{2}[3]{*}{Model}& \multicolumn{3}{c}{RDVS\cite{mou2023salient}} & \multicolumn{3}{|c}{DVsial\cite{li2024dvsod}}\\
			\cmidrule{2-7}
			&$S_\alpha\uparrow$ & $F_{\beta}^{\textrm{max}}\uparrow$ & $M\downarrow$ & $S_\alpha\uparrow$ & $F_{\beta}^{\textrm{max}}\uparrow$ & $M\downarrow$
			\\
			\midrule
			1. baseline (without MSAM)& 0.869 & 0.817 & 0.030 & 0.849  & 0.850&  0.039  \\
			
			2. +W & 0.868  & 0.819 & 0.031 & 0.847 & 0.848 &  0.041  \\

            3. +W+H & 0.871  & 0.820 & 0.029 & 0.850 & 0.849 &  0.039  \\

            4. +W+H+S & 0.873  & 0.822 & 0.030 & 0.853 & \textbf{0.852} &  \textbf{0.038}  \\
			
			5. +W+H+S+C (with MSAM)&\textbf{0.874} & \textbf{0.823} & \textbf{0.028} & \textbf{0.854} & 0.851 & \textbf{0.038} \\
			\toprule[1.5pt]
	\end{tabular}}%
	\label{Ablation4}%
    \vspace{-0.4cm}
\end{table}%

\begin{table}[t!]
	\centering
	\footnotesize
	\renewcommand\arraystretch{0.9}
	\caption{Ablation study of different fusion modules. The best results are shown in bold}\vspace{-8pt}
	\setlength{\tabcolsep}{1mm}{
		\begin{tabular}{l|ccc|ccc}
			\toprule[1.5pt]
            \multirow{2}[3]{*}{Module}& \multicolumn{3}{c}{RDVS\cite{mou2023salient}} & \multicolumn{3}{|c}{DVsial\cite{li2024dvsod}}\\
			\cmidrule{2-7}
			&$S_\alpha\uparrow$ & $F_{\beta}^{\textrm{max}}\uparrow$ & $M\downarrow$ & $S_\alpha\uparrow$ & $F_{\beta}^{\textrm{max}}\uparrow$ & $M\downarrow$
			\\
			\midrule
			1. MFA (SPNet \cite{zhou2021specificity})& 0.865 & 0.811 & 0.033 & 0.849  & 0.842&  0.042  \\

                2. MGA (MGAN \cite{li2019motion}) & 0.864  & 0.813 & 0.034 & 0.852 & 0.844 &  0.041  \\
			
			3. RFM (DCTNet+ \cite{mou2023salient}) & 0.867  & 0.815 & 0.031 & 0.849 & 0.849 &  0.040  \\

                4. CAM (CPNet \cite{hu2024cross}) & 0.870  & 0.817 & \textbf{0.028} & 0.851 & 0.847 &  0.039  \\
			
			5. MSAM (Ours) &\textbf{0.874} & \textbf{0.823} & \textbf{0.028} & \textbf{0.854} & \textbf{0.851} & \textbf{0.038} \\
			\toprule[1.5pt]
	\end{tabular}}%
	\label{Ablation5}%
    \vspace{-0.4cm}
\end{table}%

\subsection{Failure Cases}\label{sec46}
\begin{figure}
	\centering
	\includegraphics[width=0.48\textwidth]{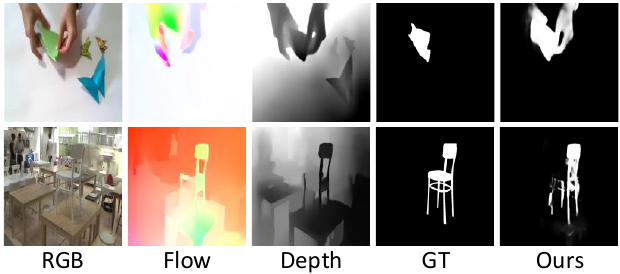}
    \vspace{-0.3cm}
	\caption{\small {Failure cases\label{fig_failure}}}
    \vspace{-0.4cm}
\end{figure}

Although our SMFNet achieves encouraging performance in RGB-D VSOD, it encounters difficulties in providing correct judgment when confronted with some extreme cases. Fig. \ref{fig_failure} illustrates some failure cases of SMFNet: (a) In the $1^{st}$ row, optical flow and depth highlight the same non-salient regions. As a result, the features we fuse through the PSF strategy will still contain this part of interference from optical flow or depth, causing some false-positive results in final prediction; (b) In the $2^{nd}$ row, RGB, optical flow and depth do not present clear edges, which results in blurred boundary of the detected saliency map. Note that case (a) also easily confuses existing methods as shown in Table \ref{tab_sota}, so it is worth exploring this case in the future. As for case (b), it can be possibly improved by using some edge refinement strategies \cite{wu2019stacked, zhao2019egnet}.

\section{CONCLUSION} 
\label{sec:conclusion}
In this paper, we propose a novel selective cross-modal fusion framework (SMFNet) to unleash the potential of incorporating optical flow and depth in RGB-D VSOD. Central to SMFNet is a pixel-level selective fusion strategy (PSF), which is proposed to selectively fuse the most valuable features of optical flow and depth based on their actual contributions. PSF consists of two key components: the generation of a spatial weight map and a pseudo-supervisory algorithm. The spatial weight map is used for the weighted fusion of optical flow and depth, while the pseudo-supervisory algorithm generates pseudo ground truth to supervise the spatial weight map during training, in order to guarantee its efficacy and correctness. Subsequently, we propose a multi-dimensional selective attention module (MSAM) to integrate the fused features derived from PSF with the remaining RGB modality at multiple dimensions, thus effectively enhancing feature representation.  Extensive experiments conducted on RDVS, DVisal, and also five VSOD datasets equipped with synthetic depth maps demonstrate the superiority of SMFNet. We make the benchmark results on RDVS and DVisal publicly available, aiming to inspire further works for RGB-D VSOD in the future.

\bibliographystyle{IEEEtran}
\bibliography{main}

\end{document}